\documentclass{article}
\usepackage{spconf,amsmath,graphicx}
\usepackage{url}
\usepackage{amssymb,amsfonts}
\usepackage{algorithmic}
\usepackage{epsfig}
\usepackage{textcomp}
\usepackage{xcolor}
\usepackage{pifont}
\usepackage{booktabs}
\usepackage{multirow} 
\usepackage{makecell}
\usepackage{colortbl} 
\usepackage{xcolor}
\usepackage{float}
\usepackage{amsmath}
\usepackage{enumitem}

\usepackage[colorlinks=true, linkcolor=black, citecolor=black, urlcolor=black, pdfborder={0 0 0}]{hyperref}
\usepackage{arydshln}
\usepackage{soul}

\def\pparagraph#1{\par{\bf #1}~}
\title{GTA-Crime: A Synthetic Dataset and Generation Framework for Fatal Violence Detection with Adversarial Snippet-Level Domain Adaptation}
\name{Seongho Kim\textsuperscript{*1} \qquad Sejong Ryu\textsuperscript{*2} \qquad Hyoukjun You\textsuperscript{*3}\thanks{\textsuperscript{*}Equal contribution.} \qquad Je Hyeong Hong\textsuperscript{\textdagger,2,3}\thanks{\textsuperscript{\textdagger} Corresponding author}}
\address{
Department of Electrical Engineering\textsuperscript{1}, Electronic Engineering\textsuperscript{2}, AI Semiconductor Engineering\textsuperscript{3}  \\
Hanyang University, Seoul, Korea
}
\begin{document}
\maketitle
\begin{abstract}
Recent advancements in video anomaly detection (VAD) have enabled identification of various criminal activities in surveillance videos, 
but detecting fatal incidents such as shootings and stabbings remains difficult due to their rarity and ethical issues in data collection.
Recognizing this limitation, we introduce GTA-Crime, a fatal video anomaly dataset and generation framework using Grand Theft Auto 5 (GTA5).
Our dataset contains fatal situations such as shootings and stabbings, captured from CCTV multiview perspectives under diverse conditions including action types, weather, time of day, and viewpoints.
To address the rarity of such scenarios, we also release a framework for generating these types of videos.
Additionally, we propose a snippet-level domain adaptation strategy using Wasserstein adversarial training to bridge the gap between synthetic GTA-Crime features and real-world features like UCF-Crime.
Experimental results validate our GTA-Crime dataset and demonstrate that incorporating GTA-Crime with our domain adaptation strategy consistently enhances real world fatal violence detection accuracy.
Our dataset and the data generation framework are publicly available at https://github.com/ta-ho/GTA-Crime.
\end{abstract}


\section{Introduction}
\label{sec:intro}
\vspace{-3mm}
In the modern society, the widespread use of surveillance cameras has become essential for public safety and crime prevention. 
However, these systems generate massive volumes of video data, exceeding the capacity of human operators to monitor effectively. 
This gap often delays responses to critical events and underscores the urgent need for robust automated methods.
Additionally, the rise in high-profile violent incidents, such as assassination attempts on political figures, has increased public anxiety and highlighted the need for an automatic reliable violence detection systems to address potential threats swiftly.

To address the aforementioned goal of automated violence detection, Sultani et al.~\cite{sultani2018real} pioneered the data-driven approach to VAD by introducing  the weakly-supervised video anomaly detection (WVAD) methodology and the UCF-Crime dataset.
Since WVAD operates under the assumption that only video-level annotations are available without existence of frame-level labels, the recent advancement of WVAD methods~\cite{rtfm2021weakly,zhou2023dual,joo2023cliptsa} ignited release of several new VAD datasets leveraging the advantage of requiring video-level labels only.

Among the newly proposed VAD datasets includes XD-Violence~\cite{xd2020not}, CCTV-Fights~\cite{perez2019detection}, GTAVEvent~\cite{montulet2021densely}, and UBnormal~\cite{acsintoae2022ubnormal} as briefly summarized in Table~\ref{tab:dataset_comparison}.
While these datasets have made strides in addressing challenges such as acquiring fatal violence scenarios~\cite{sultani2018real,xd2020not} or incorporating diverse CCTV viewpoints~\cite{montulet2021densely, acsintoae2022ubnormal}, very few manage to encompass both aspects simultaneously.
Consequently, constructing an automated, labor-free framework for generating videos with fatal violence emerges as the primary motivation for this work.

\begin{figure}[t]
    \centering
        \includegraphics[width=\columnwidth]{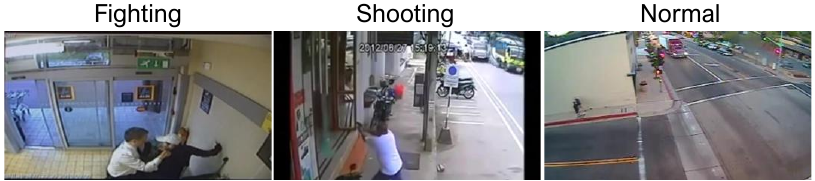}

        \includegraphics[width=\columnwidth]{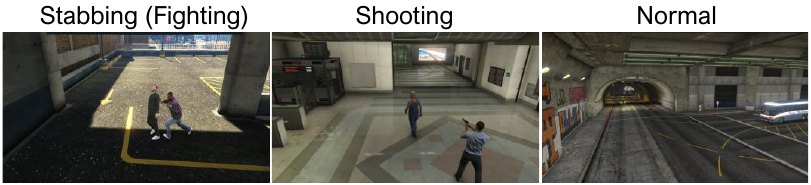}
    \vspace{-8mm}
    \caption{
    Exemplary visualization of video frames from real (UCF-Crime~\cite{sultani2018real}, top row) and our synthetic (GTA-Crime, bottom row) datasets.
    GTA-Crime comprises fatal violence clips acquired from similar environments and camera viewpoints to real CCTV footage in the UCF-Crime dataset.
    }
    \label{fig:crime_comparison}
    \vspace{-4mm}
\end{figure}

\begin{table*}[t]
\centering
\small
\caption{
    A list of known VAD datasets. 
    \# fatal videos indicates the number of videos depicting life-threatening scenarios caused by human interactions, while Data generation reflects the ability to create additional data instances.
    \textbf{$\triangle$} indicates a mixture of CCTV and non-CCTV views, while multiview refers to videos capturing the same incident from multiple perspectives.
}
\label{tab:dataset_comparison}
\vspace{1mm}
\begin{tabular}{lcccccc}  
\Xhline{1.2pt}
\textbf{Dataset} & \textbf{Fatal violence} & \textbf{\# fatal videos} & \textbf{Data generation} & \textbf{CCTV view} & \textbf{Multiview} & \textbf{Resolution} \\  
\Xhline{1.2pt}
UCF-Crime \cite{sultani2018real}    & \ding{51} & 50 & \ding{55} & \ding{51} & \ding{55} & $320 \times 240$ \\
XD-Violence \cite{xd2020not}       & \ding{51} & 546 & \ding{55} & \textbf{$\triangle$} & \ding{55} & Multiple \\
CCTV-Fights \cite{perez2019detection} & \ding{55} & \ding{55} & \ding{55} & \textbf{$\triangle$} & \ding{55} & Multiple \\
GTAVEvent \cite{montulet2021densely} & \ding{55} & \ding{55} & \ding{51} & \ding{51} & \ding{55} & $2560 \times 1440$ \\
UBnormal \cite{acsintoae2022ubnormal} & \ding{55} & \ding{55} & \ding{55} & \ding{51} & \ding{55} & Multiple \\
\hline
\textbf{GTA-Crime (ours)}          & \ding{51} & 270 & \ding{51} & \ding{51} & \ding{51} & $1920 \times 1080$\\
\Xhline{1.2pt}
\end{tabular}
\vspace{-4mm}
\end{table*}

To achieve the aforementioned goal, we introduce GTA-Crime, a novel synthetic dataset focused on fatal violence, generated in the virtual environment of the GTA game with a CCTV perspective, along with a corresponding dataset generation framework.
Our framework is designed to enable a multi-view setup, providing richer contextual information and reducing ambiguity in complex scenarios~\cite{multiview}. 
It also incorporates variations in weather, locations, and times to reflect diverse real-world conditions, ensuring a comprehensive representation of potential environments.
Table \ref{tab:dataset_comparison} compares GTA-Crime with other datasets, illustrating its focus on fatal scenarios, support for CCTV multiview perspectives, and capacity to generate data instances through a proposed framework.

While GTA-Crime provides high-quality synthetic data, deploying models trained on such data directly in real-world settings often leads to suboptimal performance due to the domain gap between virtual and actual footage. 
To address this, we propose an adversarial snippet-level domain adaptation strategy that aligns synthetic and real data representations.
This approach is tailored for VAD tasks, as it aligns synthetic and real data at the snippet level using features extracted from pre-trained models commonly used in recent VAD research~\cite{rtfm2021weakly, joo2023cliptsa, chen2023mgfn, wu2024vadclip}.
To the best of our knowledge, this is the first application of snippet-level alignment with frozen pretrained snippet feature extractors.

In summary, the main contributions of our work are:
\vspace{-3mm}
\begin{itemize}[leftmargin=*,itemsep=1mm]    
    \item Construction of a new synthetic dataset focused on fatal violence, such as shootings and stabbings, captured in a CCTV multiview setting, along with the release of the corresponding data generation framework,
    \vspace{-2mm}
    \item Introduction of a simple yet effective snippet-level domain adaptation strategy that applies adversarial training to features from pretrained snippet feature extractors.
    \vspace{-2mm}
\end{itemize}

\vspace{-3mm}
\section{Related work}
\vspace{-2mm}
\subsection{Video anomaly detection datasets}
\vspace{-2mm}

Datasets for video anomaly detection (VAD) can be broadly divided into two main categories: real-world and synthetically generated. 
Among the most widely used real-world datasets is UCF-Crime~\cite{sultani2018real}, which comprises 1,900 surveillance videos spanning 13 anomaly categories, offering substantial diversity in real-life incidents. 
XD-Violence~\cite{xd2020not} extends the variety with 4,754 videos drawn from sources like movies and sports, covering 6 major anomaly types, while CCTV-Fights~\cite{perez2019detection} narrows the scope to 1,000 videos exclusively focused on fighting scenes, typically captured by security cameras.
Real-world datasets capture genuine scenarios, but collecting severe incidents (e.g., shootings or stabbings) is challenging due to ethical concerns and their rarity.

To overcome the scarcity of high-risk incidents in real data, researchers introduced synthetic datasets to simulate extreme scenarios safely.
GTAVEvent~\cite{montulet2021densely} leverages the GTA5 game for large-scale anomaly simulations, while UBnormal~\cite{acsintoae2022ubnormal} combines animated characters with real backgrounds, creating 543 videos across 22 anomaly categories.

Although these synthetic resources mitigate some ethical and rarity issues, they often lack enough examples of life-threatening situations.
Furthermore, the mismatch between virtual and real domains often calls for additional strategies to ensure robust performance in practical applications.

\vspace{-2mm}
\begin{figure*}
    \centering
    \centerline{\includegraphics[width=0.95\textwidth]{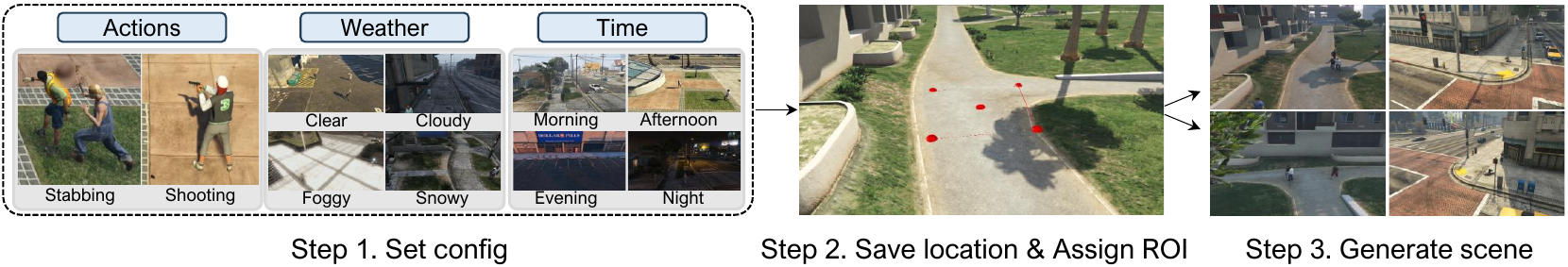}}
    \vspace{-4mm}
    
    \caption{
        An overview of video data generation pipeline for the GTA-Crime dataset.
        First, parameters such as the type of fatal violence, weather condition, and time of day are randomly assigned. 
        Second, assign a region of interest (ROI) on the game map. 
        Finally, video clips are generated from two distinct CCTV-like viewpoints to enhance dataset diversity.
    }
    \label{fig:enter-label2}
\vspace{-4mm}
\end{figure*}

\vspace{-1mm}
\subsection{Domain adaptation}
\vspace{-1mm}
The domain gap between synthetic and real data has long been recognized, prompting efforts to bridge this disparity~\cite{videoGAN, snippetDA}.
Traditional domain adaptation approaches often target high-dimensional image or video-level data, but these methods come with significant computational costs~\cite{videoGAN}. 
Alternatively, snippet-level adaptation using video encoders \cite{snippetDA} has been explored.
However, these methods typically require modifying encoders, whereas nearly all recent VAD models rely on frozen pre-trained encoders to leverage their generalization capabilities.
Adopting methods like \cite{snippetDA} for VAD necessitates altering these encoders, potentially leading to incompatibilities with state-of-the-art VAD models.
This approach also increases reliance on training data quality, risking degraded performance with low quality VAD datasets.

\vspace{-1mm}
\subsection{Weakly-supervised video anomaly detection (WVAD)}
\vspace{-1mm}
Various VAD methodologies have been proposed to develop automatic and reliable violence detection systems, with most advancements driven by WVAD~\cite{sultani2018real,rtfm2021weakly, zhou2023dual}.
Sultani et al.~\cite{sultani2018real} introduced WVAD with a multiple instance learning (MIL) framework that predicts frame-level anomalies from video-level annotations by assuming higher anomaly bag scores than normal bag scores.
Building on this work, Tian et al.~\cite{rtfm2021weakly} applied dilated convolutions and self-attention mechanisms while proposing a feature magnitude-based loss to more effectively separate normal and anomalous frames.
Zhou et al.~\cite{zhou2023dual} enhanced these ideas by incorporating a memory bank and global-local multi-head self-attention modules, enriching the model’s embedding expressiveness.

More recentely, large-scale pre-trained encoder has integrated for improved generalization~\cite{CLIP2021learning}. 
Joo et al.~\cite{joo2023cliptsa} leveraged the CLIP image encoder to capture temporal dependencies, thereby enhancing anomaly detection in complex scenes. 
Similarly, Wu et al.~\cite{wu2024vadclip} extended this idea by employing the CLIP text encoder for fine-grained anomaly classification, illustrating how text-driven cues could be exploited to refine frame-level anomaly scores.

\begin{figure}[t]
    \centering
        \includegraphics[scale=0.4]{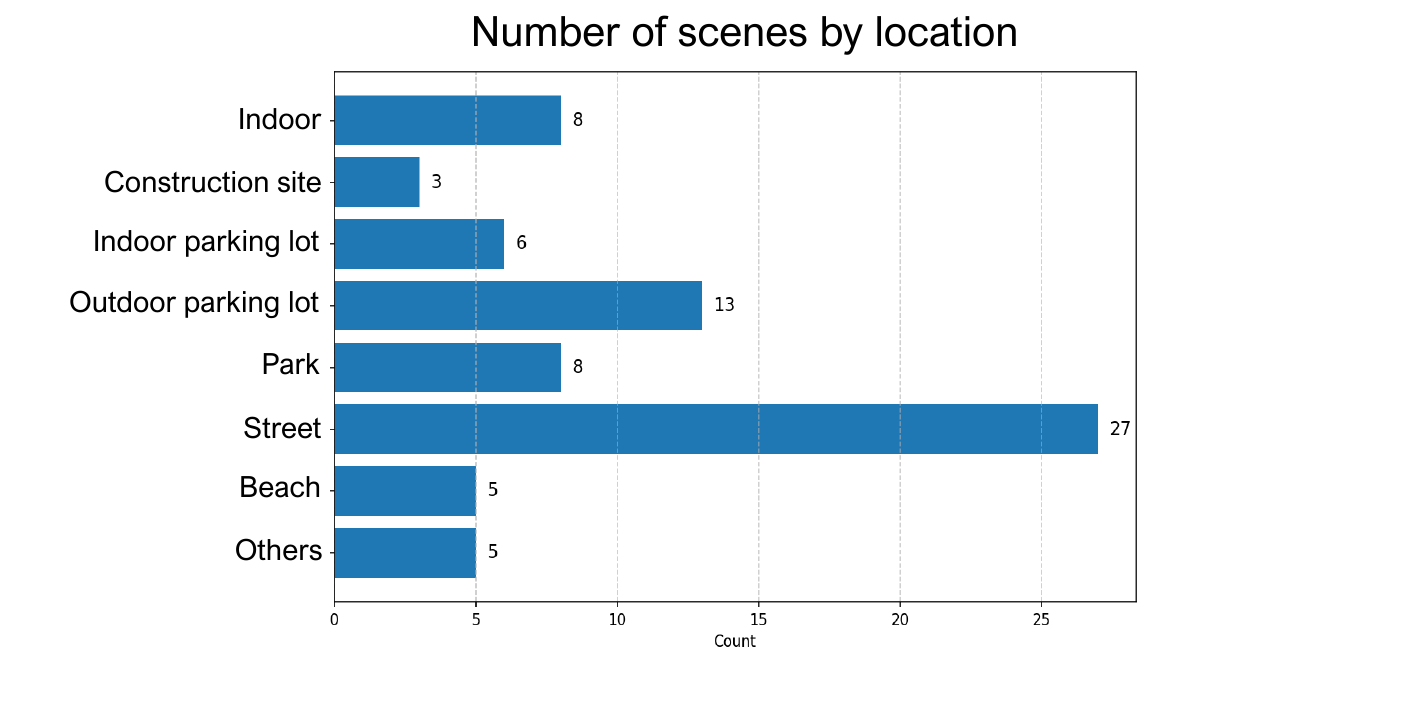}
    \vspace{-12mm}
    \caption{
    Number of scenes categorized by location. Others category includes building rooftops, swimming pools, subway station areas, gas stations, and restaurant fronts.}
    \label{fig:location}
    \vspace{-5mm}
\end{figure}

Building on the progress of WVAD methods in detecting anomalous events, we validate the effectiveness of our GTA-Crime dataset in improving fatal violence detection across diverse VAD models when integrated with real-world datasets.

\begin{figure*}
    \centering
    \includegraphics[width=1.0\linewidth]{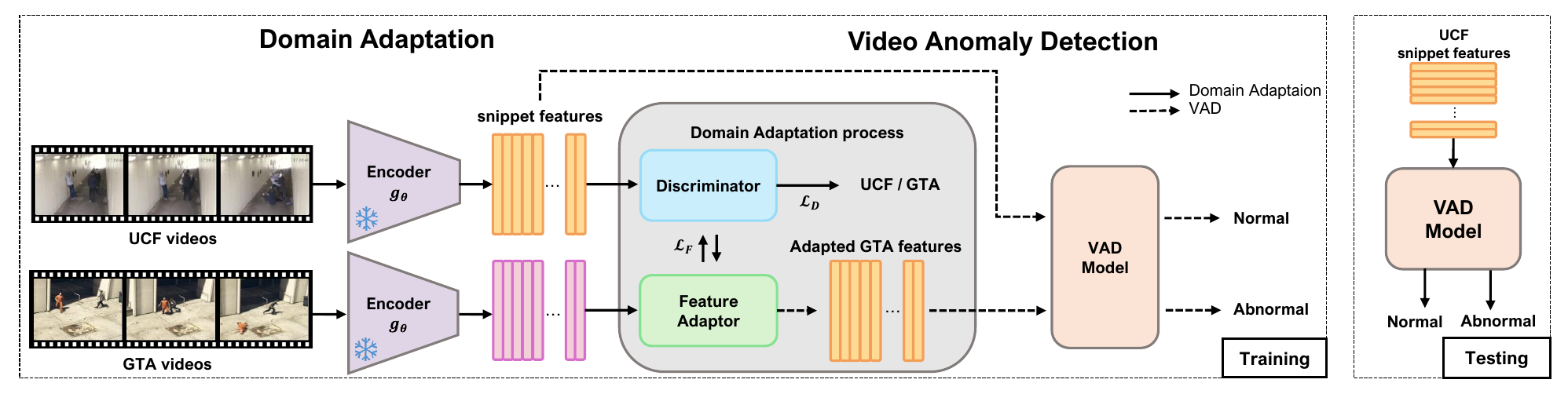} 
    \vspace{-8mm}
    \caption{
        An overview of the proposed method for improving fatal violence detection. The process begins by aligning GTA-Crime features with UCF-Crime features through the Wasserstein adversarial training. Once the features are aligned, VAD models are trained using a combination of UCF-Crime and adapted GTA-Crime features. During testing, AUC comparisons are conducted to evaluate the effectiveness of the domain adaptation strategy.}
    \label{fig:enter-label3}
    \vspace{-5mm}
\end{figure*}

\vspace{-2mm}
\section{GTA-Crime dataset: generation framework, description, and domain adaptation}
\vspace{-2mm}
\label{sec:synthetic data construction}

\subsection{Dataset generation framework} 
\vspace{-2mm}
\label{sec:data construction framework}
To generate GTA-Crime dataset, we utilize a realistic GTA environment using the Scripthook plugin to interact with the Rockstar Advanced Game Engine (RAGE).
We significantly extend the \cite{montulet2021densely} code by incorporating fatal scenarios such as stabbings and shootings, along with a CCTV multi-view setting to further enhance the dataset's diversity.
All scenes are configured to mimic real-world CCTV environments \cite{CCTVheight}, with outdoor cameras positioned at 15–20 feet and indoor cameras at 10–12 feet. 
The detailed dataset construction process is available on our Github \emph{construction.md} page.

\begin{table}[t]
    \centering
    \small
    \begin{tabular}{l|r|r|r}
        \hline
        \multicolumn{1}{c|}{Statistics} & \multicolumn{1}{c|}{Training} & \multicolumn{1}{c|}{Test} & \multicolumn{1}{c}{Total} \\
        \hline
        
        \# shooting frames & $\ 12,416\ $ & $\ 1,746\ $ & $\ 14,162\ $ \\
        \# stabbing frames & $\ 10,282\ $ & $\ 1,746\ $ & $\ 12,028\ $ \\
        \# total abnormal frames & $\ 22,698\ $ & $\ 3,492 \ $ & $\ 26,190 \ $ \\
        \# normal frames   & $\ 153,942\ $ & $\ 50,346\ $ & $\ 204,288\ $ \\
        \# abnormal minutes & $\ 12.61\ $ & $\ 1.94\ $ & $\ 14.55\ $ \\
        \# normal minutes  & $\ 85.52\ $ & $\ 27.97\ $ & $\ 113.49\ $ \\
        \# videos          & $\ 460\ $    & $\ 72\ $    & $\ 532\ $ \\
        \hline
    \end{tabular}
    \vspace{-2mm}
    \caption{Statistics of the GTA-Crime dataset. Our videos are generated at 30 FPS.}
    \label{tab:dataset_statistics}
    \vspace{-5mm}
\end{table}

The data generation pipeline shown in Fig. \ref{fig:enter-label2}, consists of three steps. 
Step 1 (Set Config): Parameters such as type of abnormal event, event timing, environmental conditions (e.g., weather, time of day) are predefined in a configuration file. 
Step 2 (Save Location \& Assign ROI): 75 GTA locations are selected as indoor and outdoor backgrounds with CCTV perspectives.
At each location, regions of interest (ROIs) are defined for spawning characters and initiating actions.
Step 3 (Generate Scene): Normal events (e.g., characters wandering or standing) and abnormal events (e.g., stabbings or shootings followed by fleeing) are simulated. 

Events are captured from two viewpoints, with all simulations recorded frame by frame.
Each frame takes approximately one second to generate and is automatically created based on the parameters set in Step 1. 
The generated frames are then sequentially compiled into videos.

\vspace{-3mm}
\subsection{Dataset description}
\vspace{-2mm}
The GTA-Crime dataset, created using the framework in Section \ref{sec:data construction framework}, includes 532 videos: 262 normal and 270 abnormal (124 stabbing, 146 shooting).
The normal videos depict people walking in everyday scenarios across 75 locations, while the abnormal videos simulate stabbing and shooting incidents in 62 and 73 locations, respectively.
Considering the diverse real-world settings of abnormal events~\cite{sultani2018real}, a wide range of locations was included in the dataset creation. 
The categorization of these settings is illustrated in Figure~\ref{fig:location}.

Each video consists of 384 frames, with abnormal events occurring between frames 192 and 288. 
The videos are captured from a CCTV viewpoint \cite{CCTVheight}, with a duration of 13 seconds at 30 fps and a resolution of $1920 \times 1080$, reflecting recent advancements in CCTV technology \cite{resolution}. 
To ensure accurate and comprehensive annotations, the type and timing of abnormal events were predetermined using a configuration file during the data generation process.
As a result, the dataset includes annotations at both video and frame levels, enabling precise temporal and spatial labeling of each event.
We release 532 videos, with the dataset generation framework in Section \ref{sec:data construction framework} allowing for the creation of additional videos.

Table~\ref{tab:dataset_statistics} presents several statistics about the GTA-Crime dataset.
The dataset consists of 532 videos, with 460 for training and 72 for testing. The dataset contains a total of 204,288 frames, with 26,190 abnormal frames and 178,098 normal frames. 
The total duration of abnormal events spans 14.55 minutes, while normal activities account for 113.49 minutes.
The overall length of our dataset is 128 minutes.

\vspace{-2mm}
\subsection{Adversarial Snippet-level Domain Adaptation}
\vspace{-1mm}
\label{sec:feature_level_domain_adaptation}

To address the inherent differences between synthetic and real-world videos \cite{videoGAN}, we propose a simple snippet-level domain adaptation strategy leveraging adversarial training with Wasserstein Gradient Penalty (WGAN-GP) \cite{WGAN}. 
The objective is to train a feature adaptor capable of transforming input GTA-Crime feature into corresponding UCF-Crime feature.

The domain adaptation process consists of two key components: a feature adaptor and a discriminator.
The feature adaptor is trained to map the GTA feature representation, $\mathcal{X}_s$, to align with the distribution of the UCF feature representation, $\mathcal{X}_t$. 
Concurrently, the discriminator is adversarially trained to distinguish between the adapted GTA features and the real UCF features. 
This adversarial setup ensures that the feature adaptor minimizes the domain gap by iteratively aligning the GTA feature representation.

The feature adaptor loss is defined as:
\vspace{-1mm}
\begin{equation}
\mathcal{L}_{F} = -\mathbb{E}[D(G(\mathcal{X}_s))],
\end{equation}
\vspace{-2mm}
and the discriminator loss is defined as:
\begin{equation}
\begin{aligned}
\mathcal{L}_{D} = & \ \mathbb{E}[D(\mathcal{X}_t)] - \mathbb{E}[D(G(\mathcal{X}_s))] \\
& + \lambda_{\text{GP}} \mathbb{E}_{\hat{\mathcal{X}}} \left[ \left( \|\nabla_{\hat{\mathcal{X}}} D(\hat{\mathcal{X}})\|_2 - 1 \right)^2 \right],
\end{aligned}
\end{equation}
\vspace{-4mm}

where \( G \) represents the feature adaptor, \( D \) denotes the discriminator, and \( \mathbb{E}[\cdot] \) computes the mean over all snippets in the batch, approximating the expectation over the respective data distributions \(( \mathcal{X}_s \), \( \mathcal{X}_t \), \( \hat{\mathcal{X}}) \) as applicable. Here, \( \hat{\mathcal{X}} \) represents interpolated samples between the source domain (\(\mathcal{X}_s\)) and the target domain (\(\mathcal{X}_t\)).
The gradient penalty term \( \lambda_{\text{GP}} \) stabilizes the adversarial training process.

In this process, we conducted class-wise domain adaptation by separately aligning features from each class—Stabbing (Fighting), Shooting, and Normal—between the GTA-Crime and UCF-Crime datasets.
Specifically, GTA Stabbing features were adapted to UCF Fighting features, GTA Shooting features to UCF Shooting, and GTA Normal features to UCF Normal. 
This class-wise alignment ensures that the domain adaptation is more precise, as features from corresponding classes are adapted to each other.

\vspace{-2mm}
\section{Experimental results}
\vspace{-2mm}
Our experiments have been designed to
\vspace{-2mm}
\begin{enumerate}
    \item validate the standalone utility of the GTA-Crime dataset through independent experiments.
    \vspace{-3mm}
    \item  evaluate the contribution of GTA-Crime snippets adapted via domain adaptation (DA) to fatal violence detection in UCF-Crime under three experimental settings:
    \vspace{-3mm}
    \begin{enumerate}
        \item UCF-Crime only,
        \vspace{-1mm}
        \item UCF-Crime with GTA-Crime (w/o DA), and
        \vspace{-1mm}
        \item UCF-Crime with GTA-Crime (w/ DA).
    \end{enumerate}
\end{enumerate}
\vspace{-2mm}

The first experiment demonstrates the validity of the GTA-Crime dataset by showing comparable results to the UCF-Crime benchmark.
Additionally, applying DA to align GTA snippets with UCF, we confirm improvements in fatal violence detection, validating our dataset and DA strategy.

\pparagraph{Implementation details.}
The evaluation involves three methods using I3D features~\cite{rtfm2021weakly, zhou2023dual, chen2023mgfn} and two utilizing CLIP features~\cite{joo2023cliptsa, wu2024vadclip}, both of which extract features from snippets comprising 16 consecutive frames.
I3D~\cite{I3D2017quo} generates 1024-dimensional features, while CLIP (ViT-B/16)~\cite{CLIP2021learning} produces 512-dimensional features with pre-trained frozen encoders.

Following feature extraction, DA is applied to align GTA features with UCF features, which are then combined with the UCF features.
The VAD models are trained and evaluated on an NVIDIA RTX A6000 GPU, using equal numbers of GTA and UCF features for balanced training. 
We used the default settings for each VAD model, including batch size, learning rate, and optimizer, and measured performance using area under the ROC curve (AUC) averaged over three seeds.
Additional training details are in the supplementary material~\cite{supmat}.

\vspace{-2mm}
\subsection{Analysis of GTA-Crime dataset}
\vspace{-2mm}
Table~\ref{tab:gta_crime_results} presents fatal violence detection results on our GTA-Crime dataset, using frame-level labels to enable experiments under both supervised and weakly-supervised settings.

\pparagraph{Experiment setting.}
The dataset was initially generated with 262 normal, 124 stabbing, and 146 shooting videos.
Each category was then randomly split into training and testing sets, resulting in 226/36 for normal, 106/18 for stabbing, and 128/18 for shooting videos, following a similar ratio to the splits used in UCF-Crime~\cite{sultani2018real}.

For comparison with real-world data, we also define UCF3, a subset of the UCF-Crime dataset~\cite{sultani2018real}, which includes three classes—Fighting (45/5), Shooting (27/23), and Normal (72/28) videos for training/testing—to emphasize the importance of detecting high-risk behaviors. 
This subset is designed to emphasize high-risk behaviors and provide a fair evaluation baseline for domain adaptation experiments.

In the supervised setting (highlighted as light orange color) in Table~\ref{tab:gta_crime_results}, the original process of selecting the top-$k$ snippet features from abnormal videos was modified. 
Using the frame-level labels, the code was adjusted to extract features and indices of the corresponding snippets for experiments.
All experiments except for \textbf{single view} were conducted in the multiview setting, while the \textbf{single view} results represent the average of two individual views.

\pparagraph{Results.}
In Table~\ref{tab:gta_crime_results}, the supervised setting consistently outperforms the weakly-supervised setting across all experiments, highlighting the advantage of our dataset generation framework in providing hard-to-obtain frame-level labels.
In weakly-supervised setting, the \textbf{Overall} (multiview) setting outperformed the \textbf{Single view} across all VAD models, demonstrating its effectiveness in reducing ambiguity and enhancing detection with richer contextual information.
The comparable performance to UCF3 results in Table~\ref{tab:performance_comparison} supports the standalone validity of our dataset.
For ablation purposes, the \textbf{Shoot} and \textbf{Stab} columns correspond to shooting and stabbing scenarios, respectively.

\definecolor{one_class}{RGB}{255, 245, 230}
\definecolor{light_purple}{RGB}{238, 223, 251}
\begin{table}
\centering
\caption{AUC comparison of VAD algorithms in 
\sethlcolor{one_class}\hl{supervised} and 
\sethlcolor{light_purple}\hl{weakly-supervised} settings on GTA-Crime. AUC is computed per category using only the corresponding test data, and the overall AUC is not simply the average of individual AUCs.}
\label{tab:gta_crime_results}
\small
\resizebox{\columnwidth}{!}{
\begin{tabular}{l||>{\columncolor{one_class}}c|>{\columncolor{light_purple}}c>{\columncolor{light_purple}}c>{\columncolor{light_purple}}c>{\columncolor{light_purple}}c}
\Xhline{1.2pt}
\textbf{Method} & \textbf{Supervised} & \textbf{Overall} & \textbf{Single view} & \textbf{Shoot.} & \textbf{Stab.} \\
\Xhline{1.2pt}
RTFM \cite{rtfm2021weakly} & 91.64 & \textbf{84.02} & 83.69 & 78.96 & 84.05   \\
UR-DMU \cite{zhou2023dual} & 86.54 & \textbf{80.13} & 74.59 & 70.09 & 71.89   \\
MGFN \cite{chen2023mgfn} & 93.13 & \textbf{76.18} & 73.07 & 77.93 & 86.55   \\
\hline
CLIP-TSA \cite{joo2023cliptsa} & 92.10 & \textbf{84.49} & 78.25 & 86.06 & 87.09   \\
VadCLIP \cite{wu2024vadclip} & 83.39 & \textbf{71.36} & 69.71 & 78.84 & 64.35   \\
\Xhline{1.2pt}
\end{tabular}
}
\vspace{-5mm}
\end{table}

\vspace{-3mm}
\subsection{Analysis of our domain adaptation strategy}
\vspace{-2mm}
Table \ref{tab:performance_comparison} summarizes the performance of various VAD models across different dataset configurations, including experiments with and without domain adaptation (DA) using WGAN-GP~\cite{WGAN} (Ours).
For the with and without DA experiments, 124 stabbing videos, 146 shooting videos, and 262 normal activity videos from the GTA-Crime dataset were added to the UCF3 training data.
Experiments with other domain adaptation methods are provided in the supplementary material~\cite{supmat}.

\pparagraph{Results.}
Without domain adaptation, models using CLIP features outperform UCF3 only settings, likely due to their model architecture leveraging textual labels for anomaly types. 
However, models using I3D features experience a performance drop, highlighting the challenges of directly combining synthetic and real features without adaptation.

With our domain adaptation strategy, all models show performance improvements over their original configurations, including further gains for CLIP-based models compared to w/o DA.
These results demonstrate the effectiveness of our dataset and DA strategy in enhancing fatal violence detection.

\begin{table}[t]
\centering
\caption{AUC comparison of different methods with UCF3 and UCF3+GTA datasets using domain adaptation.}
\resizebox{0.95\columnwidth}{!}{
\begin{tabular}{c|l|c|c|c}
\Xhline{1.2pt}
\multirow{2}{*}{\textbf{Feature}} & \multirow{2}{*}{\textbf{Method}} & \multirow{2}{*}{\textbf{UCF3}} & \multicolumn{2}{c}{\textbf{UCF3+GTA}} \\ \cline{4-5}
 & & & \textbf{w/o DA} & \textbf{Ours} \\ \Xhline{1.2pt}
\multirow{3}{*}{I3D~\cite{I3D2017quo}} 
 & RTFM \cite{rtfm2021weakly} & \underline{85.43} & 84.98 & \textbf{87.27} \\ 
 & UR-DMU \cite{zhou2023dual} & \underline{86.04} & 81.39 & \textbf{86.47} \\ 
 & MGFN \cite{chen2023mgfn} & \underline{82.55} & 79.37 & \textbf{83.64} \\ \hline
\multirow{2}{*}{CLIP~\cite{CLIP2021learning}} 
 & CLIP-TSA \cite{joo2023cliptsa} & 78.23 & \underline{81.46} & \textbf{81.71} \\ 
 & VadCLIP \cite{wu2024vadclip} & 76.25 & \underline{77.08} & \textbf{77.11} \\ \Xhline{1.2pt}
\end{tabular}
}
\vspace{-6mm}
\label{tab:performance_comparison}
\end{table}

\vspace{-2mm}
\section{Conclusion}
\vspace{-1mm}
We have addressed the problem of fatal violence detection from CCTV footage, recognizing that incidents of brutal violence pose a growing societal threat. 
To overcome the scarcity of fatal violence data in surveillance videos, we designed a fatal-scenario generation framework, which facilitates the creation of additional training data. 
Building on this, we constructed \emph{GTA-Crime}, a synthetic dataset specifically designed to improve VAD performance for fatal scenarios. 
To further enhance its applicability, we employed a snippet-level domain adaptation strategy using Wasserstein adversarial training, aligning GTA-Crime features with those of real-world UCF-Crime. 
Experimental results confirm the validity of GTA-Crime dataset and show that the detection accuracy of fatal violence consistently improved by integrating domain-adapted GTA features.
Future work will expand the dataset with diverse scenarios and larger participant groups to enhance real-world applicability.

\vspace{-2mm}
\section{acknowledgement}
\vspace{-2mm}
This work was supported by the Institute of Information and communications Technology Planning and Evaluation (IITP) under the Artificial Intelligence Semiconductor Support Program to nurture the best talents (IITP-2025-RS-2023-00253914) grant funded by the Korean government (MSIT) and Technology Innovation Program (1415178807, Development of Industrial Intelligent Technology for Manufacturing, Process, and Logistics) funded by the Ministry of Trade, Industry \& Energy (MOTIE, Korea).

\vfill\pagebreak

\bibliographystyle{IEEEbib}
\bibliography{strings}

\end{document}